\def\year{2020}\relax
\documentclass[letterpaper]{article} 
\usepackage{aaai20}  
\usepackage{times}  
\usepackage{helvet} 
\usepackage{courier}  
\usepackage[hyphens]{url}  
\usepackage{graphicx} 
\usepackage{graphicx}  
\usepackage{amsmath}
\usepackage{amssymb}
\usepackage{subfig}
\newcommand{\citet}[1]{\citeauthor{#1}~\shortcite{#1}}

\title{Towards Scalable Multi-Domain Conversational Agents: The Schema-Guided Dialogue Dataset}
\author{Abhinav Rastogi, Xiaoxue Zang, Srinivas Sunkara, Raghav Gupta, Pranav Khaitan\\Google Research, Mountain View, California, USA\\
\{abhirast, xiaoxuez, srinivasksun, raghavgupta, pranavkhaitan\}@google.com 
}
 
\begin{document}

\maketitle

\begin{abstract}
Virtual assistants such as Google Assistant, Alexa and Siri provide a conversational interface to a large number of services and APIs spanning multiple domains. Such systems need to support an ever-increasing number of services with possibly overlapping functionality. Furthermore, some of these services have little to no training data available. Existing public datasets for task-oriented dialogue do not sufficiently capture these challenges since they cover few domains and assume a single static ontology per domain. In this work, we introduce the the Schema-Guided Dialogue (SGD) dataset, containing over 16k multi-domain conversations spanning 16 domains. Our dataset exceeds the existing task-oriented dialogue corpora in scale, while also highlighting the challenges associated with building large-scale virtual assistants. It provides a challenging testbed for a number of tasks including language understanding, slot filling, dialogue state tracking and response generation. Along the same lines, we present a schema-guided paradigm for task-oriented dialogue, in which predictions are made over a dynamic set of intents and slots, provided as input, using their natural language descriptions. This allows a single dialogue system to easily support a large number of services and facilitates simple integration of new services without requiring additional training data. Building upon the proposed paradigm, we release a model for dialogue state tracking capable of zero-shot generalization to new APIs, while remaining competitive in the regular setting.
\end{abstract}

\section{Introduction}
Virtual assistants help users accomplish tasks including but not limited to finding flights, booking restaurants and, more recently, navigating user interfaces, by providing a natural language interface to services and APIs on the web. The recent popularity of conversational interfaces and the advent of frameworks like Actions on Google and Alexa Skills, which allow developers to easily add support for new services, has resulted in a major increase in the number of application domains and individual services that assistants need to support, following the pattern of smartphone applications. 

Consequently, recent work has focused on scalable dialogue systems that can handle tasks across multiple application domains. Data-driven deep learning based approaches for multi-domain modeling have shown promise, both for end-to-end and modular systems involving dialogue state tracking and policy learning. This line of work has been facilitated by the release of multi-domain dialogue corpora such as MultiWOZ \cite{budzianowski2018multiwoz}, M2M \cite{shah2018building} and FRAMES \cite{el2017frames}.

However, existing datasets for multi-domain task-oriented dialogue do not sufficiently capture a number of challenges that arise with scaling virtual assistants in production. These assistants need to support a large \cite{kim-etal-2018-efficient}, constantly increasing number of services over a large number of domains. In comparison, existing public datasets cover few domains. Furthermore, they define a single static API per domain, whereas multiple services with overlapping functionality, but heterogeneous interfaces, exist in the real world.

 To highlight these challenges, we introduce the Schema-Guided Dialogue (SGD) dataset\footnote{The dataset has been released at github.com/google-research-datasets/dstc8-schema-guided-dialogue}, which is, to the best of our knowledge, the largest public task-oriented dialogue corpus. It exceeds existing corpora in scale, with over 16000 dialogues in the training set spanning 26 services belonging to 16 domains (more details in Table \ref{table:datasets}). Further, to adequately test the models' ability to generalize in zero-shot settings, the evaluation sets contain unseen services and domains. The dataset is designed to serve as an effective testbed for intent prediction, slot filling, state tracking and language generation, among other tasks in large-scale virtual assistants.

We also propose the schema-guided paradigm for task-oriented dialogue, advocating building a single unified dialogue model for all services and APIs. Using a service's schema as input, the model would make predictions over this dynamic set of intents and slots present in the schema. This setting enables effective sharing of knowledge among all services, by relating semantically similar concepts across APIs, and allows the model to handle unseen services and APIs. Under the proposed paradigm, we present a novel architecture for multi-domain dialogue state tracking. By using large pre-trained models like BERT \cite{devlin2019bert}, our model can generalize to unseen services and is robust to API changes, while achieving competitive results on the original and updated MultiWOZ datasets \cite{eric2019multiwoz}.

\begin{table*}[t!]
\centering
    \begin{tabular}[t]{ c | c c c c c c }

    \textbf{Metric $\downarrow$ Dataset $\rightarrow$} & \textbf{DSTC2} & \textbf{WOZ2.0} & \textbf{FRAMES} & \textbf{M2M} & \textbf{MultiWOZ} & \textbf{SGD} \\\hline
    No. of domains & 1 & 1 & 3 & 2 & 7 & \textbf{16}\\
    No. of dialogues &  1,612 & 600 & 1,369 & 1,500 & 8,438 & \textbf{16,142}\\
    Total no. of turns &  23,354 & 4,472 & 19,986 & 14,796 & 113,556 & \textbf{329,964}\\
    Avg. turns per dialogue & 14.49 & 7.45 & 14.60 & 9.86 & 13.46 & \textbf{20.44} \\
    Avg. tokens per turn & 8.54 & 11.24 & 12.60 & 8.24 & \textbf{13.13} & 9.75\\
    Total unique tokens & 986 & 2,142 & 12,043 & 1,008 & 23,689 & \textbf{30,352}\\
    No. of slots &  8 & 4 & 61 & 13 & 24 & \textbf{214}\\
    No. of slot values & 212 & 99 & 3,871 & 138 & 4,510 & \textbf{14,139}\\

    \end{tabular}
    \caption{Comparison of our SGD dataset to existing related datasets for task-oriented dialogue. Note that the numbers reported are for the training portions for all datasets except FRAMES, where the numbers for the complete dataset are reported.}
    \label{table:datasets}
\end{table*}

\section{Related Work}
Task-oriented dialogue systems have constituted an active area of research for decades. The growth of this field has been consistently fueled by the development of new datasets. Initial datasets were limited to one domain, such as ATIS \cite{hemphill1990atis} for spoken language understanding for flights. The Dialogue State Tracking Challenges \cite{williams2013dialog,henderson2014second,Henderson2014TheTD,kim2017fourth} contributed to the creation of dialogue datasets with increasing complexity. Other notable related datasets include WOZ2.0 \cite{wen2017network}, FRAMES \cite{el2017frames}, M2M \cite{shah2018building} and MultiWOZ \cite{budzianowski2018multiwoz}. These datasets have utilized a variety of data collection techniques, falling within two broad categories:
\begin{itemize}
    \item \textbf{Wizard-of-Oz} This setup \cite{kelley1984iterative} connects two crowd workers playing the roles of the user and the system. The user is provided a goal to satisfy, and the system accesses a database of entities, which it queries as per the user's preferences. WOZ2.0, FRAMES and MultiWOZ, among others, have utilized such methods.
    \item \textbf{Machine-machine Interaction} A related line of work explores simulation-based dialogue generation, where the user and system roles are simulated to generate a complete conversation flow, which can then be converted to natural language using crowd workers as done in ~\citet{shah2018building}. Such a framework may be cost-effective and error-resistant since the underlying crowd worker task is simpler, and annotations are obtained automatically.
\end{itemize}

As virtual assistants incorporate diverse domains, recent work has focused on zero-shot modeling \cite{bapna2017towards,xia2018zero,shah-etal-2019-robust}, domain adaptation and transfer learning techniques \cite{yang2017transfer,rastogi2017scalable,zhu2018concept}. Deep-learning based approaches have achieved state of the art performance on dialogue state tracking tasks. Popular approaches on small-scale datasets estimate the dialogue state as a distribution over all possible slot-values \cite{henderson2014,wen2017network,ren2018towards}  or individually score all slot-value combinations \cite{mrkvsic2017neural,zhong-etal-2018-global}. Such approaches are not practical for deployment in virtual assistants operating over real-world services having a very large and dynamic set of possible values. Addressing these concerns, approaches utilizing a dynamic vocabulary of slot values have been proposed \cite{rastogi2018multi,goel2019hyst,wu-etal-2019-transferable}.

\section{The Schema-Guided Dialogue Dataset} \label{sec:dataset}
An important goal of this work is to create a benchmark dataset highlighting the challenges associated with building large-scale virtual assistants. Table \ref{table:datasets} compares our dataset with other public datasets. Our Schema-Guided Dialogue (SGD) dataset exceeds other datasets in most of the metrics at scale. The especially larger number of domains, slots, and slot values, and the presence of multiple services per domain, are representative of these scale-related challenges. Furthermore, our evaluation sets contain many services, and consequently slots, which are not present in the training set, to help evaluate model performance on unseen services.

The 20 domains present across the train, dev and test splits are listed in Table \ref{table:domains}. We create synthetic implementations of a total of 45 services or APIs over these domains. Our simulator framework interacts with these services to generate dialogue outlines, which are a structured representation of dialogue semantics. We then used a crowd-sourcing procedure to paraphrase these outlines to natural language utterances. Our novel crowd-sourcing procedure preserves all annotations obtained from the simulator and does not require any extra annotations after dialogue collection. In this section, we describe these steps in detail and then present analyses of the collected dataset.

\begin{table}[!htb]
    \centering
    \def\arraystretch{1.23}
    \resizebox{\columnwidth}{!}{
        \begin{tabular}{ l | cc||l | cc } 
            \textbf{Domain} & \textbf{\#Intents}  & \textbf{\#Dialogs} & \textbf{Domain} & \textbf{\#Intents}  & \textbf{\#Dialogs} \\ \hline
            Alarm$^{2,3}$ & 2 (1) & 324 & Movies$^{1,2,3}$ & 5 (3) & 2339\\
            Banks$^{1,2}$ & 4 (2) & 1021 & Music$^{1,2,3}$ & 6 (3) & 1833\\
            Buses$^{1,2,3}$ & 6 (3) & 3135 & Payment$^3$ & 2 (1) & 222\\
            Calendar$^1$ & 3 (1) & 1602 & RentalCars$^{1,2,3}$ & 6 (3) & 2510\\
            Events$^{1,2,3}$ & 7 (3) & 4519 & Restaurants$^{1,2,3}$ & 4 (2) & 3218\\
            Flights$^{1,2,3}$ & 10 (4) & 3644 & RideSharing$^{1,2,3}$ & 2 (2) & 2223\\
            Homes$^{1,2,3}$ & 2 (1) & 1273  & Services$^{1,2,3}$ & 8 (4) & 2956\\
            Hotels$^{1,2,3}$ & 8 (4) & 4992  & Train$^{3}$ & 2 (1) & 350\\
            Media$^{1,2,3}$ & 6 (3) & 1656 & Travel$^{1,2,3}$ & 1 (1) & 2808\\
            Messaging$^3$ & 1 (1) & 298 & Weather$^{1,2,3}$ & 1 (1) & 1783\\
          \end{tabular}
    }
    \caption{The total number of intents (services in parentheses) and dialogues for each domain across train$^1$, dev$^2$ and test$^3$ sets. Multi-domain dialogues contribute to counts of each constituent domain. The domain `Service' includes salons, dentists, doctors etc. The `Alarm', `Messaging', `Payment' and `Train' domains are only present in the dev or test sets to test generalization to new domains.} 
    \label{table:domains}
\end{table}

\subsection{Services and APIs}
\begin{figure}[ht]
\centering
\includegraphics[width=0.36\textwidth]{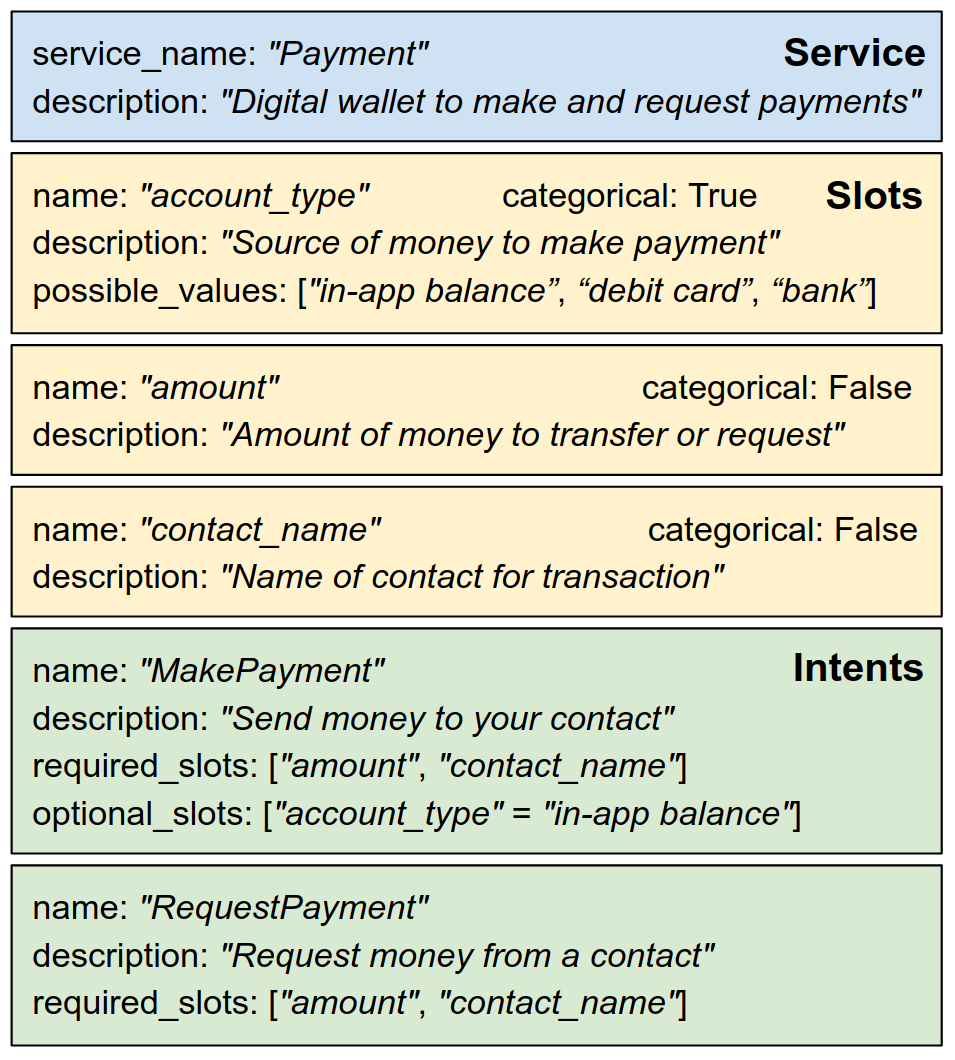}
\caption{Example schema for a digital wallet service.}
\label{fig:schema-example}
\end{figure}

We define the schema for a service as a combination of intents and slots with additional constraints, with an example in Figure~\ref{fig:schema-example}. We implement all services using a SQL engine. For constructing the underlying tables, we sample a set of entities from Freebase and obtain the values for slots defined in the schema from the appropriate attribute in Freebase. We decided to use Freebase to sample real-world entities instead of synthetic ones since entity attributes are often correlated (e.g, a restaurant's name is indicative of the cuisine served). Some slots like event dates/times and available ticket counts, which are not present in Freebase, are synthetically sampled.

To reflect the constraints present in real-world services and APIs, we impose a few other restrictions. First, our dataset does not expose the set of all possible slot values for some slots. Having such a list is impractical for slots like date or time because they have infinitely many possible values or for slots like movie or song names, for which new values are periodically added. Our dataset specifically identifies such slots as \textit{non-categorical} and does not provide a set of all possible values for these. We also ensure that the evaluation sets have a considerable fraction of slot values not present in the training set to evaluate the models in the presence of new values. Some slots like gender, number of people, day of the week etc. are defined as \textit{categorical} and we specify the set of all possible values taken by them. However, these values are not assumed to be consistent across services. E.g., different services may use (`male', `female'), (`M', `F') or (`he', `she') as possible values for gender slot.

Second, real-world services can only be invoked with a limited number of slot combinations: e.g. restaurant reservation APIs do not let the user search for restaurants by date without specifying a location. However, existing datasets simplistically allow service calls with any given combination of slot values, thus giving rise to flows unsupported by actual services or APIs. As in Figure~\ref{fig:schema-example}, the different service calls supported by a service are listed as intents. Each intent specifies a set of required slots and the system is not allowed to call this intent without specifying values for these required slots. Each intent also lists a set of optional slots with default values, which the user can override.

\subsection{Dialogue Simulator Framework}
\begin{figure}[ht]
\centering
\includegraphics[width=0.47\textwidth]{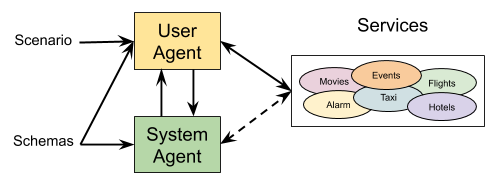}
\caption{The overall architecture of the dialogue simulation framework for generating dialogue outlines.}
\label{fig:simulator}
\end{figure}

The dialogue simulator interacts with the services to generate dialogue outlines. Figure~\ref{fig:simulator} shows the overall architecture of our dialogue simulator framework. It consists of two agents playing the roles of the user and the system. Both agents interact with each other using a finite set of actions specified through dialogue acts over a probabilistic automaton designed to capture varied dialogue trajectories.
These dialogue acts can take a slot or a slot-value pair as argument. Figure~\ref{fig:dialogue_act_distribution} shows all dialogue acts supported by the agents.

At the start of a conversation, the user agent is seeded with a scenario, which is a sequence of intents to be fulfilled. We identified over 200 distinct scenarios for the training set, each comprising up to 5 intents. For multi-domain dialogues, we also identify combinations of slots whose values may be transferred when switching intents e.g. the \textit{`address'} slot value in a restaurant service could be transferred to the \textit{`destination'} slot for a taxi service invoked right after.

The user agent then generates the dialogue acts to be output in the next turn. It may retrieve arguments i.e. slot values for some of the generated acts by accessing either the service schema or the SQL backend. The acts, combined with the respective parameters yield the corresponding user actions. Next, the system agent generates the next set of actions using a similar procedure. Unlike the user agent, however, the system agent has restricted access to the services (denoted by dashed line), e.g. it can only query the services by supplying values for all required slots for some service call. This helps us ensure that all generated flows are valid.

After an intent is fulfilled through a series of user and system actions, the user agent queries the scenario to proceed to the next intent. Alternatively, the system may suggest related intents e.g. reserving a table after searching for a restaurant. The simulator also allows for multiple intents to be active during a given turn. While we skip many implementation details for brevity, it is worth noting that we do not include any domain-specific constraints in the simulation automaton. All domain-specific constraints are encoded in the schema and scenario, allowing us to conveniently use the simulator across a wide variety of domains and services.

\subsection{Dialogue Paraphrasing}

\begin{figure}[ht]
\centering
\includegraphics[width=0.42\textwidth]{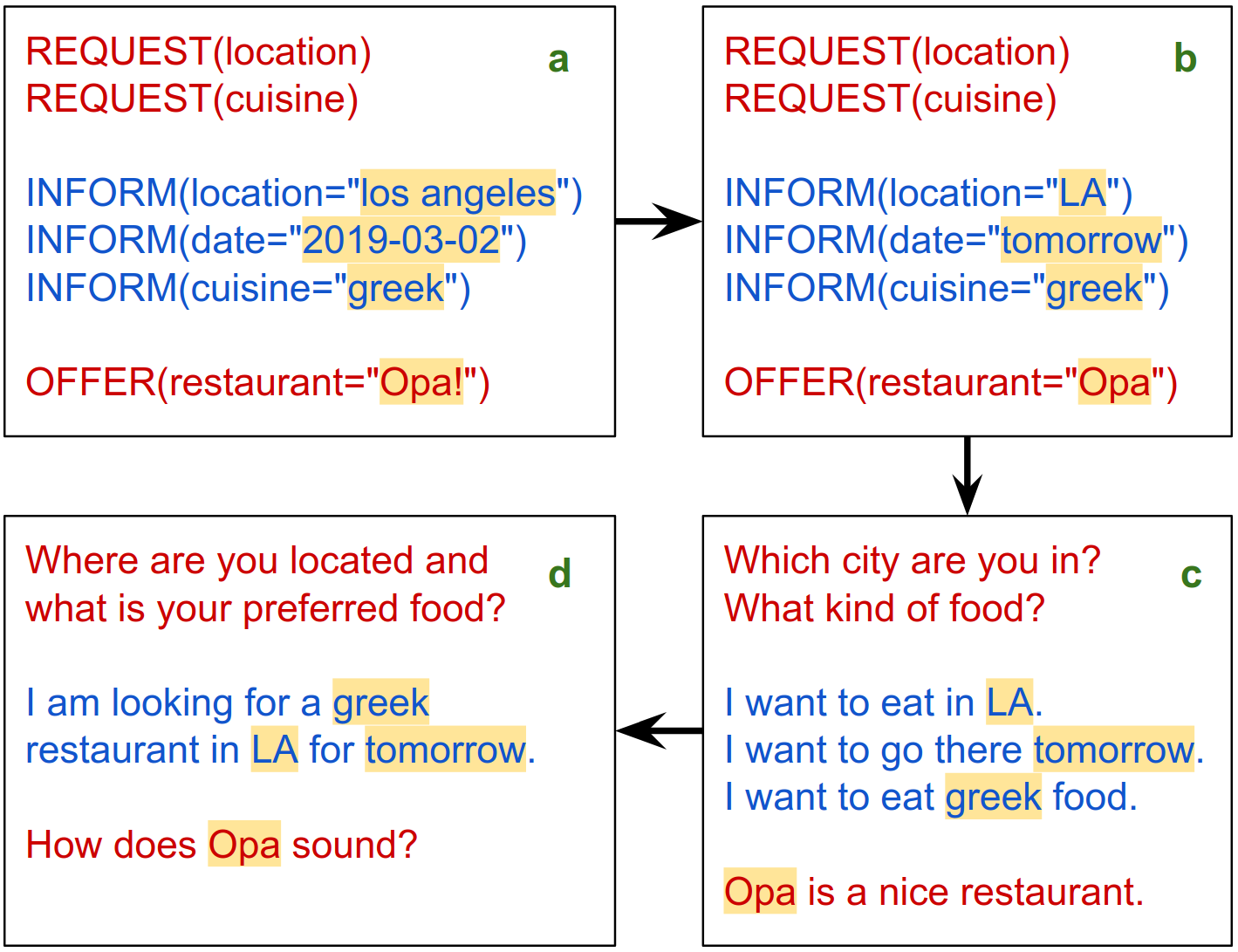}
\caption{Steps for obtaining paraphrased conversations. To increase the presence of relative dates like tomorrow, next Monday, the current date is assumed to be March 1, 2019.}
\label{fig:paraphrasing}
\end{figure}

The dialogue paraphrasing framework converts the outlines generated by the simulator into a natural conversation. Figure~\ref{fig:paraphrasing}a shows a snippet of the dialogue outline generated by the simulator, containing a sequence of user and system actions. The slot values present in these actions are in a canonical form because they obtained directly from the service. However, users may refer to these values in various different ways during the conversation, e.g., ``los angeles" may be referred to as ``LA" or ``LAX". To introduce these natural variations in the slot values, we replace different slot values with a randomly selected variation (kept consistent across user turns in a dialogue) as shown in Figure~\ref{fig:paraphrasing}b.

Next we define a set of action templates for converting each action into a utterance. A few examples of such templates are shown below. These templates are used to convert each action into a natural language utterance, and the resulting utterances for the different actions in a turn are concatenated together as shown in Figure~\ref{fig:paraphrasing}c. The dialogue transformed by these steps is then sent to the crowd workers. One crowd worker is tasked with paraphrasing all utterances of a dialogue to ensure naturalness and coherence.
\begin{align*}
\texttt{REQUEST(location)} &\rightarrow \text{Which city are you in?} \\
\texttt{INFORM(location=\$x)} &\rightarrow \text{I want to eat in \$x.} \\
\texttt{OFFER(restaurant=\$x)} &\rightarrow \text{\$x is a nice restaurant.}
\end{align*}

In our paraphrasing task, the crowd workers are instructed to exactly repeat the slot values in their paraphrases. This not only helps us verify the correctness of the paraphrases, but also lets us automatically obtain slot spans in the generated utterances by string search. This automatic slot span generation greatly reduced the annotation effort required, with little impact on dialogue naturalness, thus allowing us to collect more data with the same resources. Furthermore, it is important to note that this entire procedure preserves all other annotations obtained from the simulator including the dialogue state. Hence, no further annotation is needed.

\subsection{Dataset Analysis}

With over 16000 dialogues in the training set, the Schema-Guided Dialogue dataset is the largest publicly available annotated task-oriented dialogue dataset. The annotations include the active intents and dialogue states for each user utterance and the system actions for every system utterance. We have a few other annotations like the user actions but we withhold them from the public release. These annotations enable our dataset to be used as benchmark for tasks like intent detection, dialogue state tracking, imitation learning of dialogue policy, dialogue act to text generation etc. The schemas contain semantic information about the APIs and the constituent intents and slots, in the form of natural language descriptions and other details (example in Figure \ref{fig:schema-example}).

\begin{figure}
    \centering
    \subfloat[Histogram of lengths of training set dialogues.]{  \includegraphics[width=0.95\columnwidth]{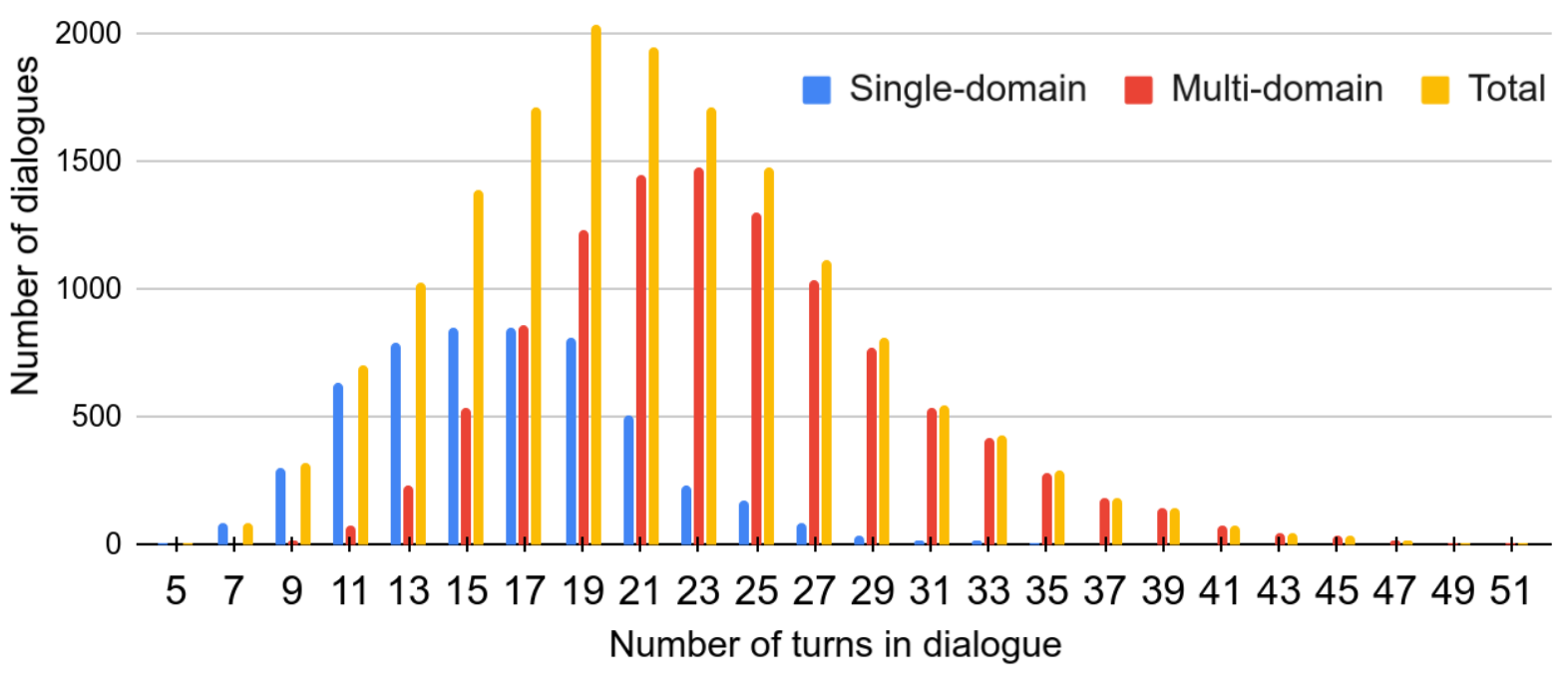} \label{fig:dialogue_lengths}} \qquad
\subfloat[Distribution of dialogue acts in training set.]{\includegraphics[width=0.95\columnwidth]{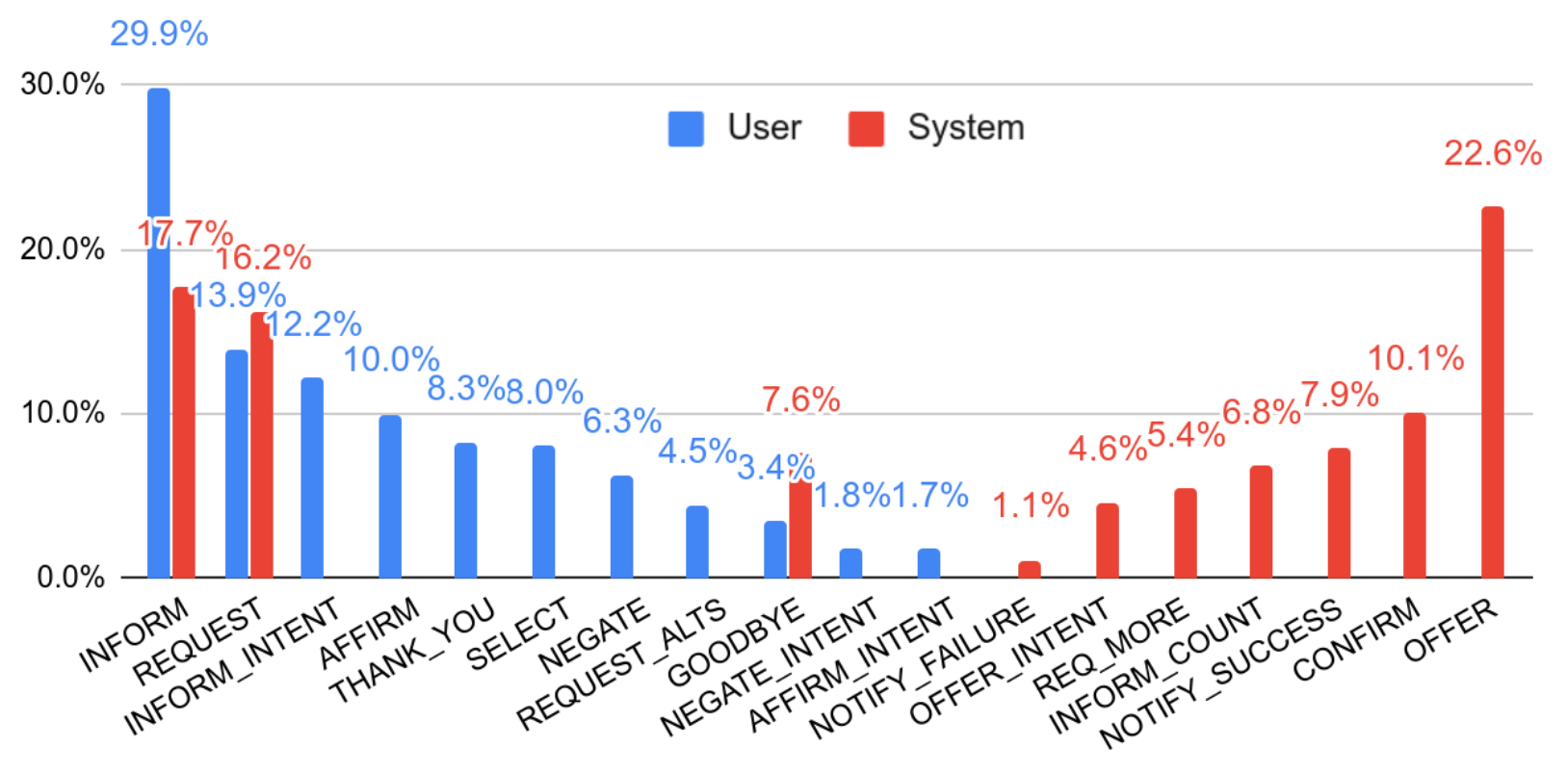}
\label{fig:dialogue_act_distribution}}

  \caption{Detailed statistics of the SGD dataset.}
\end{figure}

 The single-domain dialogues in our dataset contain an average of 15.3 turns, whereas the multi-domain ones contain 23 turns on an average. These numbers are also reflected in Figure \ref{fig:dialogue_lengths} showing the histogram of dialogue lengths on the training set.  Table \ref{table:domains} shows the distribution of dialogues across the different domains. We note that distribution of dialogues across the domains and services covered is largely balanced, with the exception domains which are not present in the training set. Figure \ref{fig:dialogue_act_distribution} shows the frequency of dialogue acts contained in the dataset. Note that all dialogue acts except \texttt{INFORM}, \texttt{REQUEST} and \texttt{GOODBYE} are specific to either the user or the system.

\section{The Schema-Guided Approach}
\begin{figure*}[t!]
\centering
\includegraphics[width=0.92\textwidth]{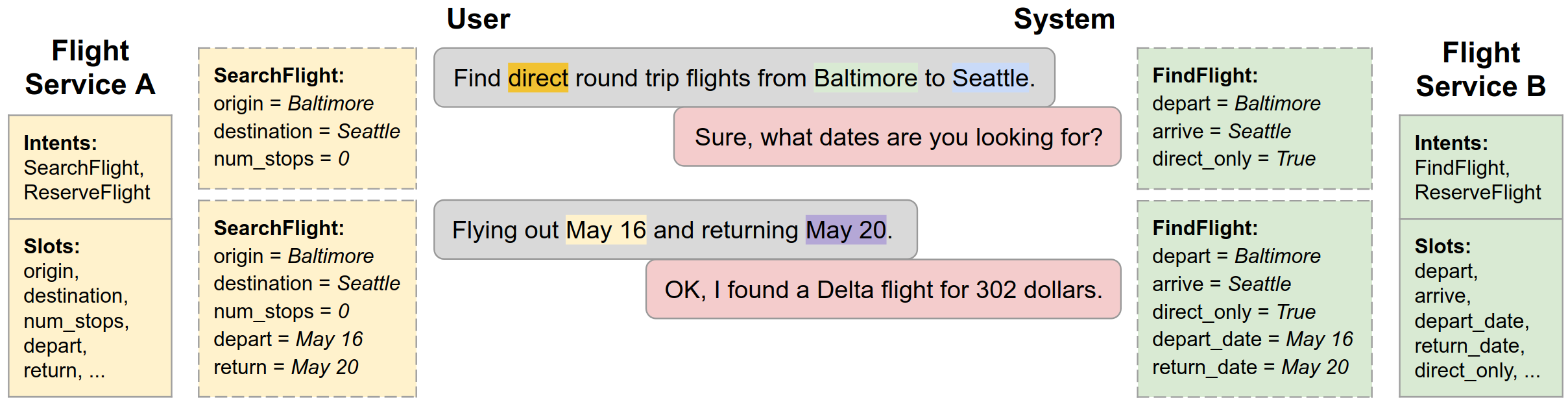}
\caption{The predicted dialogue state (shown with dashed edges) for the first two user turns for an example dialogue, showing the active intent and slot assignments, with two related annotation schemas. Note that the dialogue state representation is conditioned on the schema under consideration, which is provided as input, as are the user and system utterances.}
\label{fig:schema-guided-example}
\end{figure*}

Virtual assistants aim to support a large number of services available on the web. One possible approach is to define a large unified schema for the assistant, to which different service providers can integrate with. However, it is difficult to come up with a common schema covering all use cases. Having a common schema also complicates integration of tail services with limited developer support. We propose the schema-guided approach as an alternative to allow easy integration of new services and APIs.

Under our proposed approach, each service provides a schema listing the supported slots and intents along with their natural language descriptions (Figure \ref{fig:schema-example} shows an example). These descriptions are used to obtain a semantic representation of these schema elements. The assistant employs a single unified model containing no domain or service specific parameters to make predictions conditioned on these schema elements. For example, Figure \ref{fig:schema-guided-example} shows how dialogue state representation for the same dialogue can vary for two different services. Here, the departure and arrival cities are captured by analogously functioning but differently named slots in both schemas. Furthermore, values for the \textit{number\_stops} and \textit{direct\_only} slots highlight idiosyncrasies between services interpreting the same concept.

Using a single model facilitates representation and transfer of common knowledge across related services. Since the model utilizes semantic representation of schema elements as input, it can interface with unseen services or APIs on which it has not been trained. It is also robust to changes like addition of new intents or slots to the service.


\section{Zero-Shot Dialogue State Tracking}
Models in the schema-guided setting can condition on the pertinent services' schemas using descriptions of intents and slots. These models, however, also need access to representations for potentially unseen inputs from new services. Recent pretrained models like ELMo \cite{peters2018deep} and BERT \cite{devlin2019bert} can help, since they are trained on very large corpora. Building upon these, we present a simple prototype model for zero-shot schema-guided DST.

\subsection{Model}
We use a single model\footnote{Our model code is available at github.com/google-research/google-research/tree/master/schema\_guided\_dst}, shared among all services and domains, to make these predictions. We first encode all the intents, slots and slot values for categorical slots present in the schema into an embedded representation. Since different schemas can have differing numbers of intents or slots, predictions are made over dynamic sets of schema elements by conditioning them on the corresponding schema embeddings. This is in contrast to existing models which make predictions over a static schema and are hence unable to share knowledge across domains and services. They are also not robust to changes in schema and require the model to be retrained with new annotated data upon addition of a new intent, slot, or in some cases, a slot value to a service.

\subsubsection{Schema Embedding}
\begin{figure}[ht]
\centering
\includegraphics[width=0.86\columnwidth]{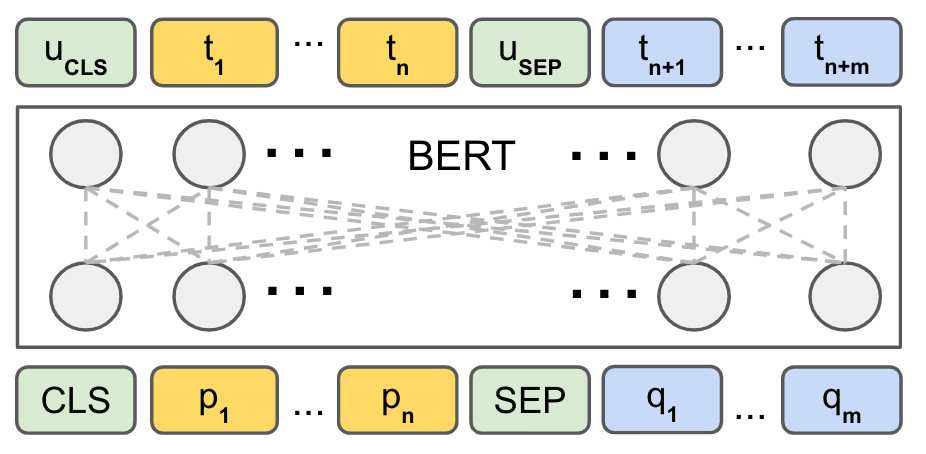}
\caption{BERT encoder, taking in two sequences $p$ and $q$ as input and outputs an embedded sequence pair representation $\mathbf{u}_{\texttt{CLS}}$ and token level representations $\{\mathbf{t}_1 \cdots \mathbf{t}_{n+m}\}$. We use BERT to obtain schema element embeddings and encode system and user utterances for dialogue state tracking.}
\label{fig:bert-model}
\end{figure}

This component obtains the embedded representations of intents, slots and categorical slot values in each service schema. Table \ref{table:schema-embedding} shows the sequence pairs used for embedding each schema element. These sequence pairs are fed to a pretrained BERT encoder shown in Figure~\ref{fig:bert-model} and the output $\mathbf{u}_{\texttt{CLS}}$ is used as the schema embedding.

\begin{table}[ht]
\centering
\begin{tabular}{l|c|c}
                & \textbf{Sequence 1}             &\textbf{Sequence 2}              \\ \hline
\textbf{Intent} & service description             & intent description              \\ \hline
\textbf{Slot}   & service description             & slot description                \\ \hline
\textbf{Value}  & slot description                & value                           \\
\end{tabular}
\caption{Input sequences for the pretrained BERT model to obtain embeddings of different schema elements.}
\label{table:schema-embedding}
\end{table}

For a given service with $I$ intents and $S$ slots, let $\{\mathbf{i}_j\}$, ${1 \leq j \leq I}$ and $\{\mathbf{s}_j\}$, ${1 \leq j \leq S}$ be the embeddings of all intents and slots respectively. As a special case, we let $\{\mathbf{s}^n_j\}$, ${1 \leq j \leq N \leq S}$ denote the embeddings for the $N$ non-categorical slots in the service. Also, let $\{\textbf{v}_j^k\}$, $1 \leq j \leq V^k$ denote the embeddings for all possible values taken by the $k^{\text{th}}$ categorical slot, $1 \leq k \leq C$, with $C$ being the number of categorical slots and $N + C = S$. All these embeddings are collectively called schema embeddings.

\subsubsection{Utterance Encoding}
Like ~\citet{chao2019bert}, we use BERT to encode the user utterance and the preceding system utterance to obtain utterance pair embedding $\mathbf{u} = \mathbf{u}_{\texttt{CLS}}$ and token level representations $\mathbf{t}_1, \mathbf{t}_2 \cdots \mathbf{t}_M$, $M$ being the total number of tokens in the two utterances. The utterance and schema embeddings are used together to obtain model predictions using a set of projections (defined below).

\subsubsection{Projection} Let $\mathbf{x}, \mathbf{y} \in \mathbb{R}^d$. For a task $K$, we define $\mathbf{l} = \mathcal{F}_K(\mathbf{x}, \mathbf{y}, p)$ as a projection transforming $\mathbf{x}$ and $\mathbf{y}$ into the vector $\mathbf{l} \in \mathbb{R}^p$ using Equations \ref{eqn:projection1}-\ref{eqn:projection3}. Here, $\mathbf{h_1},\mathbf{h_2} \in \mathbb{R}^d$, $W^K_i$ and $b^K_i$ for $1 \leq i \leq 3$ are trainable parameters of suitable dimensions and $A$ is the activation function. We use $\texttt{gelu}$ \cite{hendrycks2016gaussian} activation as in BERT.
\begin{align}
    \mathbf{h_1} &= A(W^K_1 \mathbf{x} + b^K_1) \label{eqn:projection1} \\
    \mathbf{h_2} &= A(W^K_2 (\mathbf{y} \oplus \mathbf{h_1}) + b^K_2)  \\
    \mathbf{l} &= W^K_3 \mathbf{h_2} + b^K_3 \label{eqn:projection3}
\end{align}

\subsubsection{Active Intent}
For a given service, the active intent denotes the intent requested by the user and currently being fulfilled by the system. It takes the value ``NONE" if no intent for the service is currently being processed. Let $\mathbf{i}_0$ be a trainable parameter in $\mathbb{R}^d$ for the ``NONE" intent. We define the intent network as below.
\begin{align}
l^{j}_{\text{int}} = \mathcal{F}_{\text{int}}(\mathbf{u}, \mathbf{i}_j, 1) ,\,\,\,  0 \leq j \leq I
\end{align}

The logits $l^{j}_{\text{int}}$ are normalized using softmax to yield a distribution over all $I$ intents and the ``NONE" intent. During inference, we predict the highest probability intent as active.

\subsubsection{Requested Slots}
These are the slots whose values are requested by the user in the current utterance. Projection $\mathcal{F}_{\text{req}}$ predicts logit $l^j_{\text{req}}$ for the $j^{\text{th}}$ slot. Obtained logits are normalized using sigmoid to get a score in $[0,1]$. During inference, all slots with $\text{score} > 0.5$ are predicted as requested.
\begin{align}
l^{j}_{\text{req}} = \mathcal{F}_{\text{req}}(\mathbf{u}, \mathbf{s}_j, 1) ,\,\,\,  1 \leq j \leq S
\end{align}

\subsubsection{User Goal}
We define the user goal as the user constraints specified over the dialogue context till the current user utterance. Instead of predicting the entire user goal after each user utterance, we predict the difference between the user goal for the current turn and preceding user turn. During inference, the predicted user goal updates are accumulated to yield the predicted user goal. We predict the user goal updates in two stages. First, for each slot, a distribution of size 3 denoting the slot status and taking values \texttt{none}, \texttt{dontcare} and \texttt{active} is obtained by normalizing the logits obtained in equation \ref{eqn:status} using softmax. If the status of a slot is predicted to be \texttt{none}, its assigned value is assumed to be unchanged. If the prediction is \texttt{dontcare}, then the special \texttt{dontcare} value is assigned to it. Otherwise, a slot value is predicted and assigned to it in the second stage.
\begin{align}
\label{eqn:status}
\mathbf{l}^j_{\text{status}} &= \mathcal{F}_{\text{status}}(\mathbf{u}, \mathbf{s}_j, 3) ,\,\,\,  1 \leq j \leq S \\
\label{eqn:value}
l^{j, k}_{\text{value}} &= \mathcal{F}_{\text{value}}(\mathbf{u}, \mathbf{v}^k_j, 1) ,\,\,\,  1 \leq j \leq V^k, 1 \leq k \leq C \\
\label{eqn:start}
l^{j, k}_{\text{start}} &= \mathcal{F}_{\text{start}}(\mathbf{t}_k, \mathbf{s}^n_j, 1) ,\,\,\,  1 \leq j \leq N, 1 \leq k \leq M \\
\label{eqn:end}
l^{j, k}_{\text{end}} &= \mathcal{F}_{\text{end}}(\mathbf{t}_k, \mathbf{s}^n_j, 1) ,\,\,\,  1 \leq j \leq N, 1 \leq k \leq M
\end{align}

In the second stage, equation \ref{eqn:value} is used to obtain a logit for each value taken by each categorical slot. Logits for a given categorical slot are normalized using softmax to get a distribution over all possible values. The value with the maximum mass is assigned to the slot. For each non-categorical slot, logits obtained using equations \ref{eqn:start} and \ref{eqn:end} are normalized using softmax to yield two distributions over all tokens. These two distributions respectively correspond to the start and end index of the span corresponding to the slot. The indices $p \leq q$ maximizing $start[p] + end[q]$ are predicted to be the span boundary and the corresponding value is assigned to the slot.

\subsection{Evaluation}
We consider the following metrics for evaluation of the dialogue state tracking task: 

\begin{enumerate}
    \item \textbf{Active Intent Accuracy:} The fraction of user turns for which the active intent has been correctly predicted.
    \item \textbf{Requested Slot F1:} The macro-averaged F1 score for requested slots over all eligible turns. Turns with no requested slots in ground truth and predictions are skipped.
    \item \textbf{Average Goal Accuracy:} For each turn, we predict a single value for each slot present in the dialogue state. The slots which have a non-empty assignment in the ground truth dialogue state are considered for accuracy. This is the average accuracy of predicting the value of a slot correctly. A fuzzy matching score is used for non-categorical slots to reward partial matches with the ground truth.
    \item \textbf{Joint Goal Accuracy:} This is the average accuracy of predicting \textit{all} slot assignments for a turn correctly. For non-categorical slots a fuzzy matching score is used.
\end{enumerate}

\subsubsection{Performance on other datasets} We evaluate our model on public datasets WOZ2.0 and MultiWOZ 2.1 \cite{eric2019multiwoz}. As results in Table \ref{table:metrics} show, our model performs competitively on these datasets. In these experiments, we omit the use of fuzzy matching scores and use exact match while calculating the goal accuracies to keep our numbers comparable with other works. Furthermore, for the MultiWOZ 2.1 dataset, we also trained a model incorporating pointer-generator style copying for non-categorical slots, similar to \citet{wu-etal-2019-transferable}, giving us a joint goal accuracy of \textbf{0.489}, exceeding the best-known result of 0.456 as reported in ~\citet{eric2019multiwoz}. We omit the details of this model since it is not the main focus of this work.

\subsubsection{Performance on SGD} 
\begin{figure}[ht]
\centering
\includegraphics[width=0.47\textwidth]{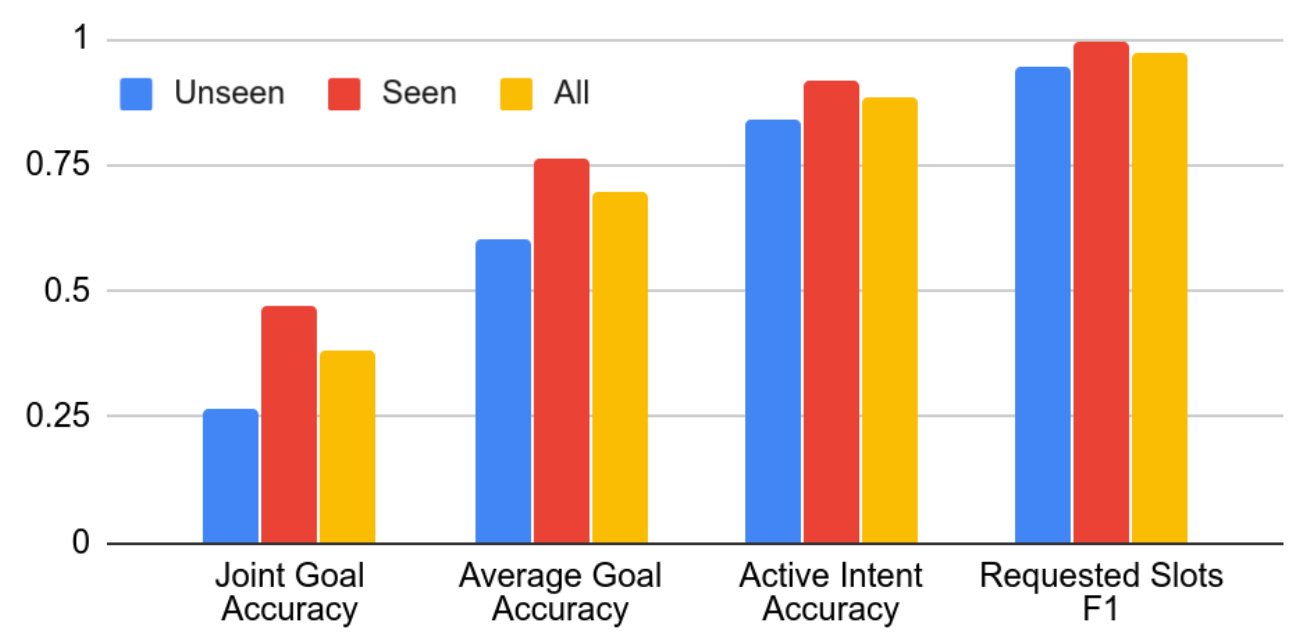}
\caption{Performance of the model on all services, services seen in training data, services not seen in training data.}
\label{fig:all_metrics}
\end{figure}

The model performs well for Active Intent Accuracy and Requested Slots F1 across both seen and unseen services, shown in Table \ref{table:metrics}. For joint goal and average goal accuracy, the model performs better on seen services compared to unseen ones (Figure~\ref{fig:all_metrics}). The main reason for this performance difference is a significantly higher OOV rate for slot values of unseen services.

\subsubsection{Performance on different domains (SGD)} The model performance also varies across various domains. The performance for the different domains is shown in Table~\ref{table:metrics_by_domain}. We observe that one of the major factors affecting the performance across domains is still the presence of the service in the training data (seen services). In most cases, the performance can be observed to degrade for domains with more unseen services. Among the unseen services, those in the `RentalCars' and `Buses' domain, have a very high OOV rate for slot values leading to worse performance. A low joint goal accuracy and high average goal accuracy for these two domains indicates a possible skew between the performance of different slots. Among seen services, `RideSharing' domain also exhibits poor performance, since it possesses the largest number of the possible slot values across the dataset. We also notice that for categorical slots, with similar slot values (e.g. ``Psychologist" and ``Psychiatrist"), there is a very weak signal for the model to distinguish between the different classes, resulting in inferior performance.

\begin{table}[ht] 
\centering
    \setlength\tabcolsep{2pt}
    \def\arraystretch{1.2}
\resizebox{\columnwidth}{!}{
\begin{tabular}{c|c|c|c|c}
\textbf{Dataset} & \textbf{Active Int Acc} &  \textbf{Req Slot F1} & \textbf{Avg GA} & \textbf{Joint GA} \\ \hline 
WOZ2.0 & N.A. & 0.970 & 0.920 & 0.810\\ \hline
MultiWOZ 2.1 & N.A. & N.A. & 0.875 & 0.434 \\ \hline\hline
SGD-S & 0.885 & 0.956 & 0.684 & 0.356 \\ \hline
SGD-All & 0.906 & 0.965 & 0.560 & 0.254 \\ \hline
\end{tabular}
}
\caption{Model performance on test sets of the respective datasets. SGD-Single model is trained on single-domain dialogues only whereas SGD-All model is trained on the entire training set. We also report results on MultiWOZ 2.1 and WOZ2.0. N.A. indicates unavailable tasks.}
\label{table:metrics}
\end{table}

\begin{table}[ht]
\centering

    \setlength\tabcolsep{4.1pt}
    \def\arraystretch{1.23}
    \resizebox{\columnwidth}{!}{
    \begin{tabular}{c|c|c||c|c|c}
\textbf{Domain} & \textbf{Joint GA} & \textbf{Avg GA} & \textbf{Domain} & \textbf{Joint GA} & \textbf{Avg GA} \\ \hline
RentalCars* & 0.086 & 0.480 & Restaurants* & 0.228 & 0.558 \\ \hline 
Buses* & 0.097 & 0.509 & Events* & 0.235 & 0.579 \\ \hline
Messaging* & 0.102 & 0.200 & Flights* & 0.239 & 0.659 \\ \hline
Payment* & 0.115 & 0.348 & Hotels** & 0.289 & 0.582 \\ \hline
Trains* & 0.136 & 0.635 & Movies** & 0.378 & 0.686 \\ \hline
Music* & 0.155 & 0.399 & Services** & 0.409 & 0.721 \\ \hline
RideSharing & 0.170 & 0.502 & Travel & 0.415 & 0.572 \\ \hline
Media* & 0.180 & 0.308 & Alarm* & 0.577 & 0.018 \\ \hline
Homes & 0.189 & 0.727 & Weather & 0.620 & 0.764 \\

\end{tabular}
    }

\caption{Model performance per domain (GA: goal accuracy). Domains marked with `*' are those for which the service in the test set is not present in the training set. Domains like Hotels marked with `**' has one unseen and one seen service. For other domains, the service in the test set was also seen in the training set. We see that the model generally performs better for domains containing services seen during training.}
\label{table:metrics_by_domain}
\end{table}

\section{Discussion}
It is often argued that simulation-based data collection does not yield natural dialogues or sufficient coverage, when compared to other approaches such as Wizard-of-Oz. We argue that simulation-based collection is a better alternative for collecting datasets like this owing to the factors below.

\begin{itemize}
    \item \textbf{Fewer Annotation Errors:} All annotations are automatically generated, so these errors are rare. In contrast, ~\citet{eric2019multiwoz} reported annotation errors in 40\% of turns in MultiWOZ 2.0 which utilized a Wizard-of-Oz setup. 
    \item \textbf{Simpler Task:} The crowd worker task of paraphrasing a readable utterance for each turn is simple. The error-prone annotation task requiring skilled workers is not needed. Furthermore, Wizard-of-Oz style collection requires domain specific task definitions and instructions, making the collection of a diverse dataset like ours time consuming.
    \item \textbf{Low Cost:} The simplicity of the crowd worker task and lack of an annotation task greatly cut data collection costs.
    \item \textbf{Better Coverage:} A wide variety of dialogue flows can be collected and specific usecases can be targeted.
\end{itemize}

To ensure naturalness of the generated conversations, we used the conversational flows present in other public datasets like MultiWOZ 2.0 and WOZ2.0 as a guideline while developing the dialogue simulator. It was difficult for us to conduct a side-by-side comparison with existing datasets since this is the first dataset to cover many new domains at such scale, but we plan to explore it in the future.

\section{Conclusions}
We presented the Schema-Guided Dialogue dataset to encourage scalable modeling approaches for virtual assistants. We also introduced the schema-guided paradigm for task-oriented dialogue that simplifies the integration of new services and APIs with large scale virtual assistants. Building upon this paradigm, we present a simplistic model for zero-shot dialogue state tracking achieving competitive results.

\subsubsection{Acknowledgements} The authors thank Guan-Lin Chao, Amir Fayazi and Maria Wang for their advice and assistance.

\fontsize{9.0pt}{10.0pt} \selectfont

\bibliography{aaai}

\begin{thebibliography}{}

\bibitem[\protect\citeauthoryear{Bapna \bgroup et al\mbox.\egroup
  }{2017}]{bapna2017towards}
Bapna, A.; T{\"{u}}r, G.; Hakkani{-}T{\"{u}}r, D.; and Heck, L.~P.
\newblock 2017.
\newblock Towards zero-shot frame semantic parsing for domain scaling.
\newblock In {\em Interspeech 2017, 18th Annual Conference of the International
  Speech Communication Association, Stockholm, Sweden, August 20-24, 2017}.

\bibitem[\protect\citeauthoryear{Budzianowski \bgroup et al\mbox.\egroup
  }{2018}]{budzianowski2018multiwoz}
Budzianowski, P.; Wen, T.-H.; Tseng, B.-H.; Casanueva, I.; Ultes, S.; Ramadan,
  O.; and Gasic, M.
\newblock 2018.
\newblock Multiwoz-a large-scale multi-domain wizard-of-oz dataset for
  task-oriented dialogue modelling.
\newblock In {\em Proceedings of the 2018 Conference on Empirical Methods in
  Natural Language Processing},  5016--5026.

\bibitem[\protect\citeauthoryear{Chao and Lane}{2019}]{chao2019bert}
Chao, G.-L., and Lane, I.
\newblock 2019.
\newblock {BERT-DST}: Scalable end-to-end dialogue state tracking with
  bidirectional encoder representations from transformer.
\newblock In {\em INTERSPEECH}.

\bibitem[\protect\citeauthoryear{Devlin \bgroup et al\mbox.\egroup
  }{2019}]{devlin2019bert}
Devlin, J.; Chang, M.-W.; Lee, K.; and Toutanova, K.
\newblock 2019.
\newblock Bert: Pre-training of deep bidirectional transformers for language
  understanding.
\newblock In {\em Proceedings of the 2019 Conference of the North American
  Chapter of the Association for Computational Linguistics: Human Language
  Technologies, Volume 1 (Long and Short Papers)},  4171--4186.

\bibitem[\protect\citeauthoryear{El~Asri \bgroup et al\mbox.\egroup
  }{2017}]{el2017frames}
El~Asri, L.; Schulz, H.; Sharma, S.; Zumer, J.; Harris, J.; Fine, E.; Mehrotra,
  R.; and Suleman, K.
\newblock 2017.
\newblock Frames: a corpus for adding memory to goal-oriented dialogue systems.
\newblock In {\em Proceedings of the 18th Annual SIGdial Meeting on Discourse
  and Dialogue},  207--219.

\bibitem[\protect\citeauthoryear{Eric \bgroup et al\mbox.\egroup
  }{2019}]{eric2019multiwoz}
Eric, M.; Goel, R.; Paul, S.; Sethi, A.; Agarwal, S.; Gao, S.; and Hakkani-Tur,
  D.
\newblock 2019.
\newblock Multiwoz 2.1: Multi-domain dialogue state corrections and state
  tracking baselines.
\newblock {\em arXiv preprint arXiv:1907.01669}.

\bibitem[\protect\citeauthoryear{Goel, Paul, and
  Hakkani-T{\"u}r}{2019}]{goel2019hyst}
Goel, R.; Paul, S.; and Hakkani-T{\"u}r, D.
\newblock 2019.
\newblock Hyst: A hybrid approach for flexible and accurate dialogue state
  tracking.
\newblock {\em arXiv preprint arXiv:1907.00883}.

\bibitem[\protect\citeauthoryear{Hemphill, Godfrey, and
  Doddington}{1990}]{hemphill1990atis}
Hemphill, C.~T.; Godfrey, J.~J.; and Doddington, G.~R.
\newblock 1990.
\newblock The atis spoken language systems pilot corpus.
\newblock In {\em Speech and Natural Language: Proceedings of a Workshop Held
  at Hidden Valley, Pennsylvania, June 24-27, 1990}.

\bibitem[\protect\citeauthoryear{Henderson, Thomson, and
  Williams}{2014a}]{henderson2014second}
Henderson, M.; Thomson, B.; and Williams, J.~D.
\newblock 2014a.
\newblock The second dialog state tracking challenge.
\newblock In {\em Proceedings of the 15th Annual Meeting of the Special
  Interest Group on Discourse and Dialogue (SIGDIAL)},  263--272.

\bibitem[\protect\citeauthoryear{Henderson, Thomson, and
  Williams}{2014b}]{Henderson2014TheTD}
Henderson, M.; Thomson, B.; and Williams, J.~D.
\newblock 2014b.
\newblock The third dialog state tracking challenge.
\newblock {\em 2014 IEEE Spoken Language Technology Workshop (SLT)}  324--329.

\bibitem[\protect\citeauthoryear{Henderson, Thomson, and
  Young}{2014}]{henderson2014}
Henderson, M.; Thomson, B.; and Young, S.
\newblock 2014.
\newblock Word-based dialog state tracking with recurrent neural networks.
\newblock In {\em Proceedings of the 15th Annual Meeting of the Special
  Interest Group on Discourse and Dialogue (SIGDIAL)},  292--299.

\bibitem[\protect\citeauthoryear{Hendrycks and
  Gimpel}{2016}]{hendrycks2016gaussian}
Hendrycks, D., and Gimpel, K.
\newblock 2016.
\newblock Gaussian error linear units (gelus).
\newblock {\em arXiv preprint arXiv:1606.08415}.

\bibitem[\protect\citeauthoryear{Kelley}{1984}]{kelley1984iterative}
Kelley, J.~F.
\newblock 1984.
\newblock An iterative design methodology for user-friendly natural language
  office information applications.
\newblock {\em ACM Transactions on Information Systems (TOIS)} 2(1):26--41.

\bibitem[\protect\citeauthoryear{Kim \bgroup et al\mbox.\egroup
  }{2017}]{kim2017fourth}
Kim, S.; D’Haro, L.~F.; Banchs, R.~E.; Williams, J.~D.; and Henderson, M.
\newblock 2017.
\newblock The fourth dialog state tracking challenge.
\newblock In {\em Dialogues with Social Robots}. Springer.
\newblock  435--449.

\bibitem[\protect\citeauthoryear{Kim \bgroup et al\mbox.\egroup
  }{2018}]{kim-etal-2018-efficient}
Kim, Y.-B.; Kim, D.; Kumar, A.; and Sarikaya, R.
\newblock 2018.
\newblock Efficient large-scale neural domain classification with personalized
  attention.
\newblock In {\em Proceedings of the 56th Annual Meeting of the Association for
  Computational Linguistics (Volume 1: Long Papers)},  2214--2224.
\newblock Melbourne, Australia: Association for Computational Linguistics.

\bibitem[\protect\citeauthoryear{Mrk{\v{s}}i{\'c} \bgroup et al\mbox.\egroup
  }{2017}]{mrkvsic2017neural}
Mrk{\v{s}}i{\'c}, N.; S{\'e}aghdha, D.~{\'O}.; Wen, T.-H.; Thomson, B.; and
  Young, S.
\newblock 2017.
\newblock Neural belief tracker: Data-driven dialogue state tracking.
\newblock In {\em Proceedings of the 55th Annual Meeting of the Association for
  Computational Linguistics (Volume 1: Long Papers)}, volume~1,  1777--1788.

\bibitem[\protect\citeauthoryear{Peters \bgroup et al\mbox.\egroup
  }{2018}]{peters2018deep}
Peters, M.~E.; Neumann, M.; Iyyer, M.; Gardner, M.; Clark, C.; Lee, K.; and
  Zettlemoyer, L.
\newblock 2018.
\newblock Deep contextualized word representations.
\newblock {\em arXiv preprint arXiv:1802.05365}.

\bibitem[\protect\citeauthoryear{Rastogi, Gupta, and
  Hakkani-Tur}{2018}]{rastogi2018multi}
Rastogi, A.; Gupta, R.; and Hakkani-Tur, D.
\newblock 2018.
\newblock Multi-task learning for joint language understanding and dialogue
  state tracking.
\newblock In {\em Proceedings of the 19th Annual SIGdial Meeting on Discourse
  and Dialogue},  376--384.

\bibitem[\protect\citeauthoryear{Rastogi, Hakkani-T{\"u}r, and
  Heck}{2017}]{rastogi2017scalable}
Rastogi, A.; Hakkani-T{\"u}r, D.; and Heck, L.
\newblock 2017.
\newblock Scalable multi-domain dialogue state tracking.
\newblock In {\em 2017 IEEE Automatic Speech Recognition and Understanding
  Workshop (ASRU)},  561--568.
\newblock IEEE.

\bibitem[\protect\citeauthoryear{Ren \bgroup et al\mbox.\egroup
  }{2018}]{ren2018towards}
Ren, L.; Xie, K.; Chen, L.; and Yu, K.
\newblock 2018.
\newblock Towards universal dialogue state tracking.
\newblock In {\em Proceedings of the 2018 Conference on Empirical Methods in
  Natural Language Processing},  2780--2786.

\bibitem[\protect\citeauthoryear{Shah \bgroup et al\mbox.\egroup
  }{2018}]{shah2018building}
Shah, P.; Hakkani-T{\"u}r, D.; T{\"u}r, G.; Rastogi, A.; Bapna, A.; Nayak, N.;
  and Heck, L.
\newblock 2018.
\newblock Building a conversational agent overnight with dialogue self-play.
\newblock {\em arXiv preprint arXiv:1801.04871}.

\bibitem[\protect\citeauthoryear{Shah \bgroup et al\mbox.\egroup
  }{2019}]{shah-etal-2019-robust}
Shah, D.; Gupta, R.; Fayazi, A.; and Hakkani-Tur, D.
\newblock 2019.
\newblock Robust zero-shot cross-domain slot filling with example values.
\newblock In {\em Proceedings of the 57th Annual Meeting of the Association for
  Computational Linguistics},  5484--5490.
\newblock Florence, Italy: Association for Computational Linguistics.

\bibitem[\protect\citeauthoryear{Wen \bgroup et al\mbox.\egroup
  }{2017}]{wen2017network}
Wen, T.; Vandyke, D.; Mrk{\v{s}}{\'\i}c, N.; Ga{\v{s}}{\'\i}c, M.;
  Rojas-Barahona, L.; Su, P.; Ultes, S.; and Young, S.
\newblock 2017.
\newblock A network-based end-to-end trainable task-oriented dialogue system.
\newblock In {\em 15th Conference of the European Chapter of the Association
  for Computational Linguistics, EACL 2017-Proceedings of Conference},
  volume~1,  438--449.

\bibitem[\protect\citeauthoryear{Williams \bgroup et al\mbox.\egroup
  }{2013}]{williams2013dialog}
Williams, J.; Raux, A.; Ramachandran, D.; and Black, A.
\newblock 2013.
\newblock The dialog state tracking challenge.
\newblock In {\em Proceedings of the SIGDIAL 2013 Conference},  404--413.

\bibitem[\protect\citeauthoryear{Wu \bgroup et al\mbox.\egroup
  }{2019}]{wu-etal-2019-transferable}
Wu, C.-S.; Madotto, A.; Hosseini-Asl, E.; Xiong, C.; Socher, R.; and Fung, P.
\newblock 2019.
\newblock Transferable multi-domain state generator for task-oriented dialogue
  systems.
\newblock In {\em Proceedings of the 57th Annual Meeting of the Association for
  Computational Linguistics},  808--819.
\newblock Florence, Italy: Association for Computational Linguistics.

\bibitem[\protect\citeauthoryear{Xia \bgroup et al\mbox.\egroup
  }{2018}]{xia2018zero}
Xia, C.; Zhang, C.; Yan, X.; Chang, Y.; and Yu, P.
\newblock 2018.
\newblock Zero-shot user intent detection via capsule neural networks.
\newblock In {\em Proceedings of the 2018 Conference on Empirical Methods in
  Natural Language Processing},  3090--3099.
\newblock Association for Computational Linguistics.

\bibitem[\protect\citeauthoryear{Yang, Salakhutdinov, and
  Cohen}{2017}]{yang2017transfer}
Yang, Z.; Salakhutdinov, R.; and Cohen, W.~W.
\newblock 2017.
\newblock Transfer learning for sequence tagging with hierarchical recurrent
  networks.
\newblock {\em arXiv preprint arXiv:1703.06345}.

\bibitem[\protect\citeauthoryear{Zhong, Xiong, and
  Socher}{2018}]{zhong-etal-2018-global}
Zhong, V.; Xiong, C.; and Socher, R.
\newblock 2018.
\newblock Global-locally self-attentive encoder for dialogue state tracking.
\newblock In {\em Proceedings of the 56th Annual Meeting of the Association for
  Computational Linguistics (Volume 1: Long Papers)},  1458--1467.
\newblock Melbourne, Australia: Association for Computational Linguistics.

\bibitem[\protect\citeauthoryear{Zhu and Yu}{2018}]{zhu2018concept}
Zhu, S., and Yu, K.
\newblock 2018.
\newblock Concept transfer learning for adaptive language understanding.
\newblock In {\em Proceedings of the 19th Annual SIGdial Meeting on Discourse
  and Dialogue},  391--399.

\end{thebibliography}
\bibliographystyle{aaai}

\newpage
\onecolumn
\appendix
\section{Appendix}

\renewcommand\thefigure{\thesection.\arabic{figure}}    

\setcounter{figure}{0} 

\begin{figure}[ht]
\centering
\begin{tabular}{l p{15.5cm}}
\textbf{User} & My best friend and his family are coming to visit and I want to take them out while they are here. Is there a concert playing next Friday? \\
\textbf{System} & I can look into that but let's narrow it down a bit first. What city or cities would you like me to search in? \\
\textbf{User} & I would prefer it to be fairly local so NY would be my first choice. \\
\textbf{System} & Amber Run Brooklyn appears to be playing at Warsaw. \\
\textbf{User} & I don't much care for that one, can you find me a different concert please? \\
\textbf{System} & There will also be Chris Botti playing at Sony Hall. \\
\textbf{User} & Let's try some other kind of activity. How about games, preferably baseball. \\
\textbf{System} & I was able to find 8 baseball games the first of which is the Mets Vs Cubs playing at Citi Field. \\
\textbf{User} & That sounds like something we would all enjoy very much. \\
\textbf{System} & Would you like for me to go ahead and purchase tickets for you in advance? \\
\textbf{User} & Yes, can you reserve 3 tickets for that game? \\
\textbf{System} & You want 3 tickets to the Mets Vs Cubs game taking place in New York next Friday, is that correct? \\
\textbf{User} & Yes, that's what I want. Can you give me the address of the event and the starting time as well? \\
\textbf{System} & I was able to book those tickets for you. The game will be held at 123-01 Roosevelt Avenue and it begins at 6 pm. \\
\textbf{User} & Thanks for the research and help, that is the only thing I needed for now. \\
\textbf{System} & It was my pleasure, have a great day and enjoy the game. \\
\end{tabular}

\caption{An example single domain dialogue from our dataset in the ``Events" domain.}
\label{fig:single-domain}
\end{figure}

\begin{figure}[h]
\centering
\begin{tabular}{l p{15.5cm}}
\textbf{User} &  I wanna change my hair style, help me find a salon \\
\textbf{System} &  Where are you trying to get your hair done? \\
\textbf{User} &  Let's find a place in Oakland \\
\textbf{System} &  How about 17 Jewels Salon in Oakland, I hear it's nice \\
\textbf{User} &  Umm...Doesn't sound like something I'd like, find me something else \\
\textbf{System} &  What about this other Oakland salon called Benny Adem Grooming Parlor \\
\textbf{User} &  What's their contact info and address? \\
\textbf{System} &  You can reach them at 510-227-4268 and they're at 408 14th Street \\
\textbf{User} &  Actually find me something else \\
\textbf{System} &  Bibo Salon in Oakland has great reviews \\
\textbf{User} &  Yes that' perfect \\
\textbf{System} &  Shall I book an appointment for you then? \\
\textbf{User} &  Yeah I want it done ASAP, maybe later today if possible? \\
\textbf{System} &  What time then? \\
\textbf{User} &  afternoon 12:30 would be great \\
\textbf{System} &  Confirming your 12:30 pm Bibo Salon appointment today \\
\textbf{User} &  Yes that's correct \\
\textbf{System} &  Ok glad I was able to book it \\
\textbf{User} &  Thanks that's it for now \\
\textbf{System} &  Can't wait to see your new hair! \\
\end{tabular}

\caption{An example single domain dialogue from our dataset in the ``Services" domain.}
\label{fig:single-domain}
\end{figure}

\begin{figure}[ht]
\centering
\begin{tabular}{l p{15cm}}
\textbf{User} &  I'd like to catch a movie nearby. Can you find me any movie showings? \\
\textbf{System} &  Sure, where should I look? \\
\textbf{User} &  Please look around SFO for movie showings. \\
\textbf{System} &  Okay, I found 10 movies being shown in that area that you might like. What do you think about the movies Ash Is Purest White, Captain Marvel or Citizen Kane? \\
\textbf{User} &  Citizen Kane sounds like a good one. \\
\textbf{System} &  How about buying any tickets for the movie? \\
\textbf{User} &  No thanks, not right now. I'd like to grab something to eat around there too. Can you find any good restaurants to eat at? \\
\textbf{System} &  Are you looking for a restaurant that serves a certain kind of food? \\
\textbf{User} &  Yeah, can you find restaurants that serve Punjabi food and are moderate with their price range? \\
\textbf{System} &  How about eating at a nice restaurant in San Francisco called August 1 Five? \\
\textbf{User} &  Great, that sounds perfect. When can I watch the movie you found me earlier? \\
\textbf{System} &  How about seeing it on March 2nd? \\
\textbf{User} &  Yeah, that works for me. \\
\textbf{System} &  Okay, there's 1 showing at 7:30 pm at the Vogue Theatre. \\
\textbf{User} &  Great, that works. \\
\textbf{System} &  Anything else I can help you with today? \\
\textbf{User} &  Yeah, can you book me a table at the restaurant you found for me? \\
\textbf{System} &  Sure, what time would you like to eat there? \\
\textbf{User} &  Reserve a table for me at around 5:30 pm, dinner time. \\
\textbf{System} &  Okay, please confirm that you'd like to reserve a table for 2 at August 1 Five located in San Francisco for March 2nd at 5:30 pm. \\
\textbf{User} &  Yup, that sounds right. Where exactly are they located and do you know if they play live music? \\
\textbf{System} &  Your table has been successfully booked! Their address is 524 Van Ness Avenue and unfortunately they don't have live music here. \\
\textbf{User} &  Okay, can you get me a cab to take me there? \\
\textbf{System} &  Sure, do you have a ride preference? I can get you a regular, luxury or other type of ride if you'd like. \\
\textbf{User} &  Yes, please get me a luxury ride. \\
\textbf{System} &  Okay, please confirm that you want a luxury ride with 2 seats to take you to 524 Van Ness Avenue. \\
\textbf{User} &  Yeah that sounds right to me. \\
\textbf{System} &  Okay, your cab has been booked and it's on its way. \\
\textbf{User} &  Awesome, thanks! \\
\textbf{System} &  Anything else I can do for you today? \\
\textbf{User} &  Nope, that's all I needed, thanks! \\
\textbf{System} &  No problem, enjoy your day! \\

\end{tabular}

\caption{An example multi-domain dialogue from our dataset covering ``Movies", ``Restaurants" and ``RideSharing" domains.}
\end{figure}

\begin{figure}[ht]
\centering
\begin{tabular}{l p{15cm}}
\textbf{User} &  My buddy is getting married and I need to find a bus ticket. \\
\textbf{System} &  I can help you with that. Can you give me a little more information like where you are traveling from, where you are traveling to, and when you would like to leave. \\
\textbf{User} &  I need to go from Las Vegas to Sacramento on the 7th of this month. \\
\textbf{System} &  Great, I was able to find 4 buss's matching that criteria. How does a bus with 1 transfer, leaving at 7:20 am, for \$88 sound? \\
\textbf{User} &  That sounds like just what i'm looking for. I will also be needing a rental car until the 11th of this month. \\
\textbf{System} &  Will this car be from Sacramento from March 7th. \\
\textbf{User} &  I would like to pick up the car on the 8th around two in the afternoon. \\
\textbf{System} &  The first of 5 cars I have found for you is a standard Accord, located at SMF International Airport, available for pickup on March 8th. \\
\textbf{User} &  That one sounds perfect. \\
\textbf{System} &  Would you like me to proceed with making that reservation for you? \\
\textbf{User} &  Yes, let's go ahead with that car. \\
\textbf{System} &  Before I reserve this I would like to make sure all the details I have are correct. You would like to pickup a standard vehicle from SMF International Airport on March 8th at 2 pm, and will be returning the car on March 11th. \\
\textbf{User} &  Yes, that is all correct but can you give me an estimated price too please? \\
\textbf{System} &  I was successfully able to reserve that with a total cost of \$105 for you. \\
\textbf{User} &  Great, could you also help me find an online movie to watch? \\
\textbf{System} &  What genre of movie are you looking for? \\
\textbf{User} &  I am really into detective movies right now. \\
\textbf{System} &  The 2 movies I found that you may like are Body Double and Citizen Kane. \\
\textbf{User} &  I would really like to see Body Double now if possible, I heard it's pretty good. \\
\textbf{System} &  Okay, so you would like me to begin playing the movie Body Double without subtitles now. Is that correct? \\
\textbf{User} &  Yes, that would be great. \\
\textbf{System} &  I've went ahead and started playing that movie for you. \\
\textbf{User} &  Thanks so much! \\
\textbf{System} &  Is there anything else I can help you with for now? \\
\textbf{User} &  No, that's all for now. \\
\textbf{System} &  Have a good day and enjoy your movie! \\

\end{tabular}

\caption{An example multi-domain dialogue from our dataset covering ``Buses", ``RentalCars" and ``Media" domains.}
\end{figure}

\begin{figure*}
\centering
\includegraphics[width=0.97\textwidth]{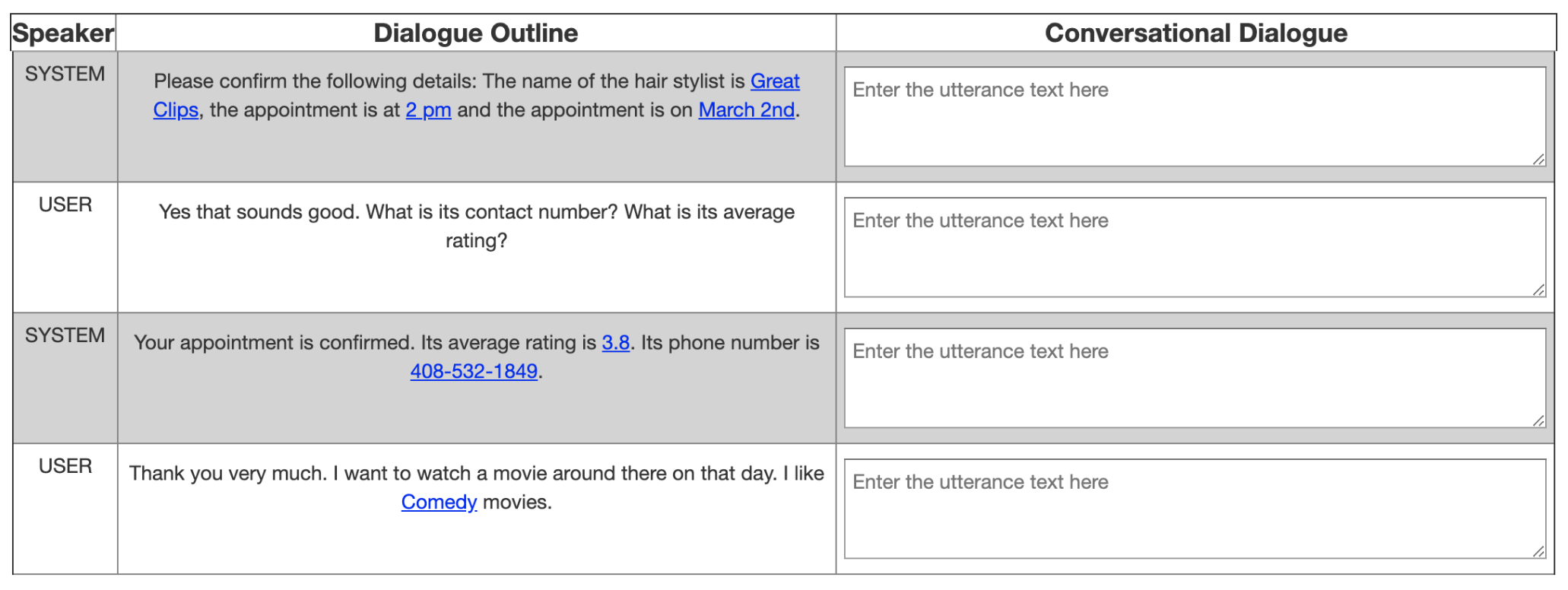}

\caption{Interface of the dialogue paraphrasing task where the crowd workers are asked to rephrase the dialogue outlines to a more natural expression. The actual interface shows the entire conversation, but only a few utterances have been shown in this figure. All non-categorical slot values are highlighted in blue. The task cannot be submitted unless all highlighted values in the outline are also present in the conversational dialogue.}
\label{fig:crowd-interface}
\end{figure*}

\end{document}